\documentclass{article}

\usepackage{arxiv}

\usepackage[utf8]{inputenc} 
\usepackage[T1]{fontenc}    
\usepackage{hyperref}       
\usepackage{url}            
\usepackage{booktabs}       
\usepackage{amsfonts}       
\usepackage{nicefrac}       
\usepackage{microtype}      
\usepackage{lipsum}
\usepackage{graphicx}
\usepackage{textcomp}
\usepackage{subfigure}
\usepackage{subcaption}
\usepackage{diagbox}
\usepackage{multirow}
\usepackage{amsmath}

\title{ELM-DeepONets: Backpropagation-Free Training of Deep Operator Networks via Extreme Learning Machines}

\author{
 Hwijae Son \\
  Department of Mathematics\\
  Konkuk University\\
  Seoul\\
  \texttt{hwijaeson@konkuk.ac.kr} \\
}

\begin{document}
\maketitle
\begin{abstract}
Deep Operator Networks (DeepONets) are among the most prominent frameworks for operator learning, grounded in the universal approximation theorem for operators. However, training DeepONets typically requires significant computational resources. To address this limitation, we propose ELM-DeepONets, an Extreme Learning Machine (ELM) framework for DeepONets that leverages the backpropagation-free nature of ELM. By reformulating DeepONet training as a least-squares problem for newly introduced parameters, the ELM-DeepONet approach significantly reduces training complexity. Validation on benchmark problems, including nonlinear ODEs and PDEs, demonstrates that the proposed method not only achieves superior accuracy but also drastically reduces computational costs. This work offers a scalable and efficient alternative for operator learning in scientific computing.  
\end{abstract}

\keywords{DeepONets \and Extreme Learning Machine \and Forward-Inverse problems}

\section{Introduction}
\label{sec:introduction}
The emergence of Physics-Informed Machine Learning (PIML) has revolutionized the intersection of computational science and artificial intelligence \cite{karniadakis2021physics}. PIML frameworks integrate physical laws, often expressed as partial differential equations (PDEs), into machine learning models to ensure that the predictions remain consistent with the underlying physics. This paradigm shift has enabled efficient and accurate solutions to complex, high-dimensional problems that were traditionally intractable with conventional numerical methods \cite{hu2024tackling, sirignano2018dgm, park2023deep}.

Physics-Informed Neural Networks (PINNs) have been at the forefront of PIML applications \cite{raissi2019physics}. By incorporating governing equations as soft constraints within the loss function, PINNs solve forward and inverse problems for PDEs without requiring large labeled datasets \cite{lu2021deepxde, cai2021physics, jo2020deep, son2024pinn}. Despite their versatility, PINNs face significant challenges, particularly in scenarios requiring frequent retraining. The need to repeatedly optimize network parameters from the initialization for varying problem instances renders PINNs computationally expensive and time-consuming.

In response to these limitations, Deep Operator Networks (DeepONets) \cite{lu2021learning}, and Neural Operators (NOs) \cite{kovachki2023neural}, have emerged as powerful alternatives. Unlike PINNs, which solve individual instances of PDEs, DeepONets and NOs learn mappings between function spaces, enabling real-time inference across a wide range of inputs. This operator learning approach eliminates the need for retraining, making it particularly suitable for applications requiring repeated evaluations, such as parameter studies and uncertainty quantification. However, the training of NOs and DeepONets involves substantial computational overhead due to the large neural network architectures and the need for expensive training data.

A brief comparison reveals distinct differences between DeepONets and NOs. While NOs, such as the Fourier Neural Operator (FNO) \cite{li2020fourier}, utilize global representations of functions via Fourier transforms \cite{li2020fourier}, DeepONets rely on branch and trunk networks to approximate operators as a linear combination of basis functions \cite{lu2021learning}. More specifically, DeepONets parameterize the target functions as functions of the input variable, so that the physics-informed training paradigm can directly be incorporated \cite{wang2021learning, cho2024physics}. Each method has unique strengths, but both share the challenge of expensive training processes, motivating the need for more efficient alternatives. 

Deep Operator Networks (DeepONets) have proven to be highly effective as function approximators, particularly in the context of operator learning. Inspired by the universal approximation theorem for operators, DeepONets approximate mappings between infinite-dimensional function spaces by decomposing the problem into two networks: a branch network that encodes the input function’s coefficients and a trunk network that evaluates the operator at target points. This structure allows DeepONets to generalize across a variety of input-output relationships, making them suitable for solving both forward and inverse problems in scientific computing.

Variants of DeepONets have been developed to enhance their approximation capabilities. \cite{prasthofer2022variable} expands DeepONets by allowing flexible input function representations, enabling robust operator learning across diverse scenarios. Fixed basis function approaches, such as the Finite Element Operator Network and Legendre Galerkin Operator Network \cite{lee2023finite, choi2023unsupervised}, incorporate predefined basis functions to improve efficiency and accuracy. These modifications leverage domain knowledge to reduce the computational burden while maintaining flexibility in approximating complex operators. Despite these advances, the training process remains a bottleneck, necessitating the exploration of alternative training methodologies.


Extreme Learning Machine (ELM) is a feed-forward neural network architecture designed for fast and efficient learning. Introduced to overcome the limitations of traditional learning algorithms, ELM operates with a Single-Layer Fully connected Networks (SLFNs) where the weights between the input and hidden layers are randomly initialized and fixed. This unique feature eliminates the need for iterative tuning of these weights, significantly reducing computational cost. The output weights are determined analytically by minimizing the training error, e.g., least squares, making ELM training extremely fast compared to conventional methods like backpropagation. Its simplicity, efficiency, and capacity for handling large-scale datasets make ELM a powerful tool in various domains, including regression, classification, and feature extraction.

The Universal Approximation Theorem for ELMs establishes that SLFNs with randomly generated hidden parameters can approximate any continuous function on a compact domain, given a sufficient number of neurons \cite{huang2006universal}. This theoretical foundation has spurred interest in applying ELMs to a variety of tasks, including classification, and regression  \cite{ding2014extreme, huang2015trends, wang2022review}. More recently, physics-informed learning. ELMs have demonstrated promising results when applied to PINN frameworks, reducing training complexity while maintaining accuracy \cite{dwivedi2020physics, liaugmented}. This success motivates the application of ELMs to operator learning, particularly for DeepONets.

In this work, we present a novel methodology ELM-DeepONets that combines the strengths of ELM and DeepONets to address the computational challenges in operator learning. Our approach leverages ELM’s efficiency to train DeepONets by solving a least squares problem, bypassing the need for expensive gradient-based optimization. The proposed ELM-DeepONets framework is validated on a diverse set of problems, including nonlinear ODEs and forward-inverse problems for benchmark PDEs. Our experiments demonstrate that the method achieves comparable accuracy to conventional DeepONet training while significantly reducing computational costs. Additionally, we highlight the potential of ELM as a lightweight and scalable alternative for operator learning in scientific computing.

The remainder of this paper is organized as follows. Section \ref{preliminary} briefly introduces important preliminaries to our method.  Section \ref{method} details the mathematical framework and implementation of the ELM-based DeepONet. Section \ref{experiments} presents the results of numerical experiments, showcasing the effectiveness of the proposed method. Finally, Section \ref{discussion} concludes with a discussion of potential extensions and future research directions.

\section{Preliminaries}\label{preliminary}
\subsection{Extreme Learning Machines}
In this subsection, we briefly outline the ELM with SLFN for the regression problem. Let $X=[x_1, x_2, \dots, x_N]^T \in \mathbb{R}^{N\times d}$ and $Y = [y_1, y_2, \dots, y_N]^T\in\mathbb{R}^{N \times 1}$ be the input data and label, respectively, where $N$ is the number of samples and $d$ is the input dimension. The network maps $x_i$ to an output using $p$ hidden neurons with randomly initialized and fixed weights $W_1 \in \mathbb{R}^{d \times p}$. The hidden layer output is computed as $H = \sigma(XW_1)$, where $\sigma(\cdot)$ is an activation function applied in an elementwise manner. The output $\widehat{Y}$ is computed by $\widehat{Y}=HW_2 = \sigma( XW_1)W_2$ where $W_2 \in \mathbb{R}^{p\times 1}$. For fixed $W_1$ and $b$, we obtain $W_2$ by least square fit to the label $Y$, i.e., $W_2 = H^\dagger Y$ where $H^\dagger$ is the Moore-Penrose pseudoinverse of $H$. \begin{figure}[t]
  \centering
  \includegraphics[width=0.4\linewidth]{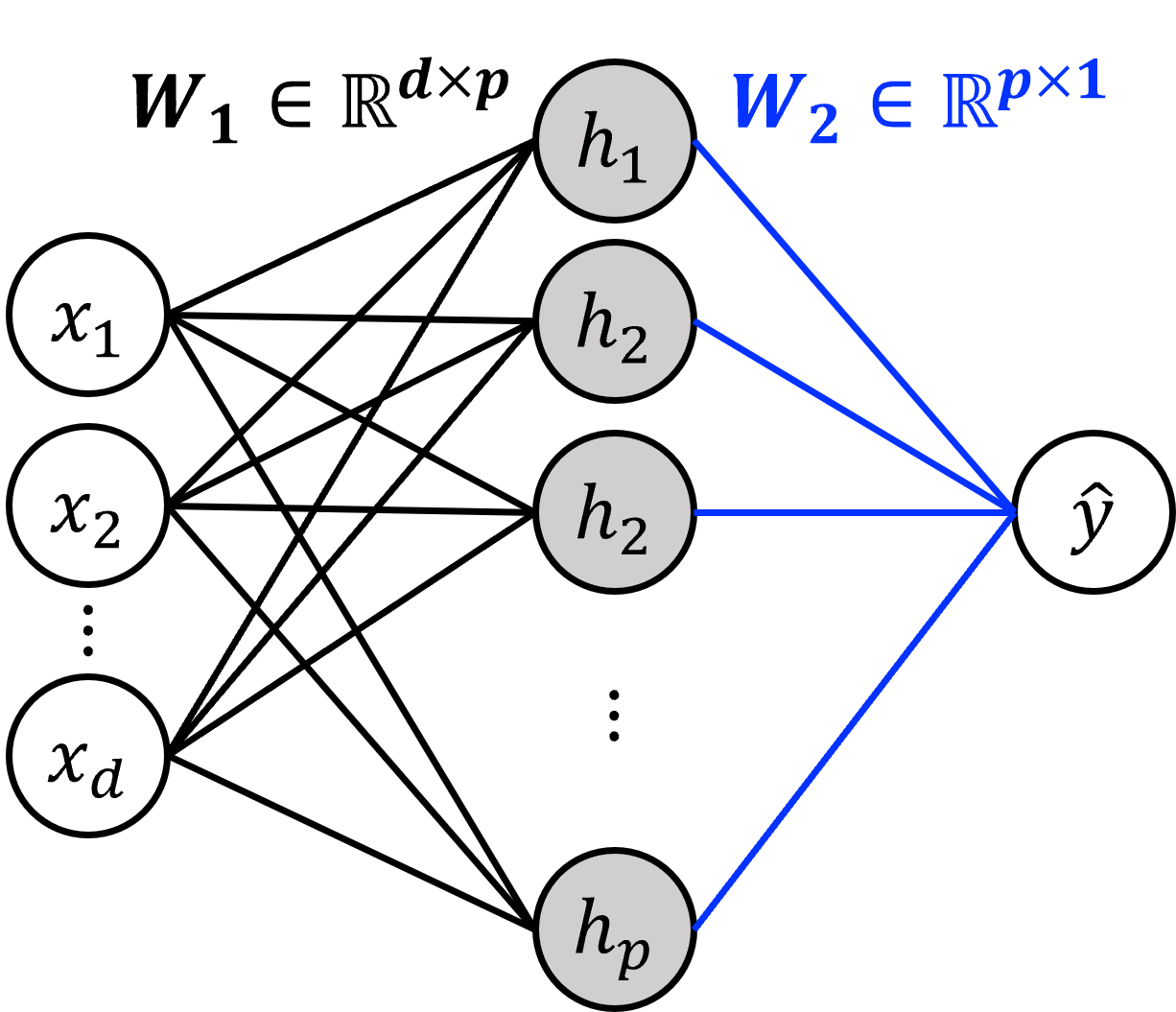}
  \caption{An illustration of ELM with SLFN. First hidden layer parameter $W_1$ is initialized randomly and fixed, while the output layer parameter $W_2$ is determined by solving a least squares problem.}
  \label{SLFN}
\end{figure}

\subsection{A brief introduction to DeepONets} DeepONet is a recently proposed operator learning framework that is usually trained in a supervised manner. The supervised dataset consists of labeled pairs of functions $\mathcal{D}=\{(u_i, F(u_i)(y_j))\}_{i,j=1}^{N,M}$, where $N$ input functions $u_i$ are characterized by $m$-dimensional vector $u_i=(u_i(x_1), u_i(x_2), \dots, u_i(x_m))$ at $m$ discrete sensor points $x_1, x_2, \dots, x_m$ and the target function $F(u)$ is evaluated at $M$ collocation points $y_1, y_2, \dots, y_M$.

\begin{figure}[t]
  \centering
  \includegraphics[width=0.7\linewidth]{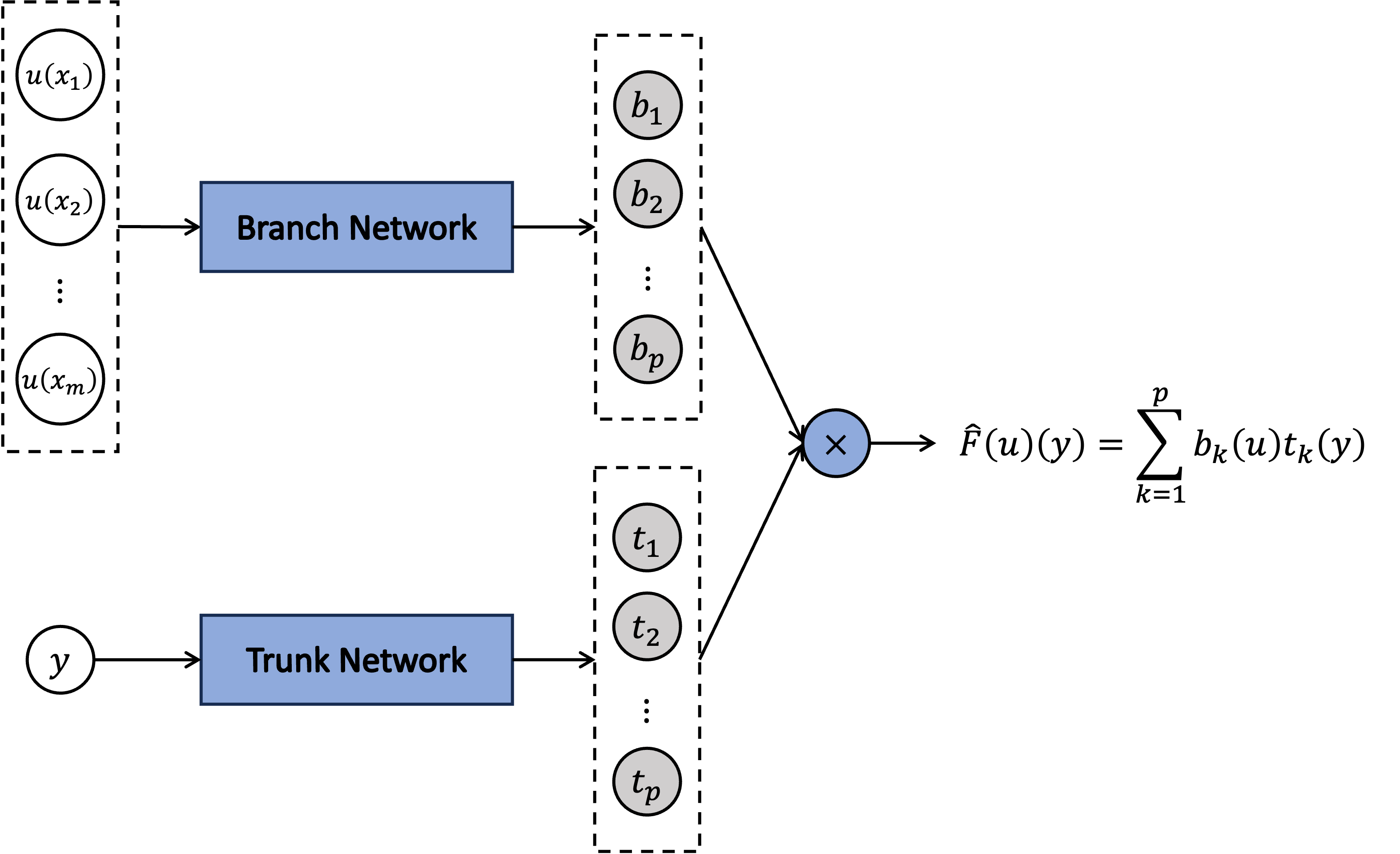}
      \caption{An illustration of DeepONet, with the picture adapted from \cite{lu2021learning}.}
   \label{deeponet}
\end{figure}

DeepONet comprises two subnetworks, the branch network and the trunk network. The branch network takes the $m$-dimensional vector $u_i=(u_i(x_1), u_i(x_2), \dots, u_i(x_m))$ and generates $p$-dimensional vector $\vec{b}=(b_1, b_2, \dots, b_p)$. On the other hand, the trunk network takes the collocation point $y$ as input to generate another $p$-dimensional vector $\vec{t}=(t_1, t_2, \dots, t_p)$. Finally, we take the inner product of $\vec{b}$ and $\vec{t}$ to generate the output $\widehat{F}(u)(y)=\sum_{k=1}^p b_k(u)t_k(y)$. Then, two subnetworks are trained to minimize the loss function defined by:

\begin{align}
    \mathcal{L}(\mathcal{D}) &= \frac{1}{2N} \sum_{i=1}^N \sum_{j=1}^M \Vert \widehat{F}(u_i)(y_j) - F(u_i)(y_j)\Vert^2 \nonumber\\
    &= \frac{1}{2N}\sum_{i=1}^N \sum_{j=1}^M \Vert \sum_{k=1}^p b_k(u_i)t_k(y_j) - F(u_i)(y_j)\Vert^2. \nonumber
\end{align} The overall architecture is illustrated in Figure \ref{deeponet}.

\section{ELM-DeepONets}\label{method}
ELM employs fixed basis functions generated by random parameters and determines the coefficients for their linear combination. Similarly, DeepONet constructs a linear combination of basis functions produced by the trunk network, with the coefficients generated by the branch network. Building on the structural similarity between ELM and DeepONet, we propose a novel framework called ELM-DeepONet. 

Recall that the output of DeepONet is expressed as:
\begin{equation}
    \widehat{F}(u)(y) = \sum_{k=1}^p b_k(u) t_k(y), \nonumber
\end{equation} where $b_k(u)$, $k=1,2,\dots,p$, are the coefficients generated from the branch network and $t_k(y)$, $k=1,2,\dots,p$, are the basis functions generated from the trunk network. To incorporate the fixed basis functions of ELM into DeepONet, we model $t_k(y)$ using a fully connected neural network with fixed parameters. This modification aligns the trunk network with the ELM philosophy by removing the need to train its parameters. In the standard ELM framework, the coefficients are determined by solving a simple least squares problem. However, in DeepONet, the coefficients $b_k(u)$ are functions of the input $u$, introducing additional complexity that makes it challenging to directly apply the ELM methodology. We address this issue by incorporating an ELM to model the coefficients. 

\begin{figure*}[t]
  \centering
  \includegraphics[width=0.9\linewidth]{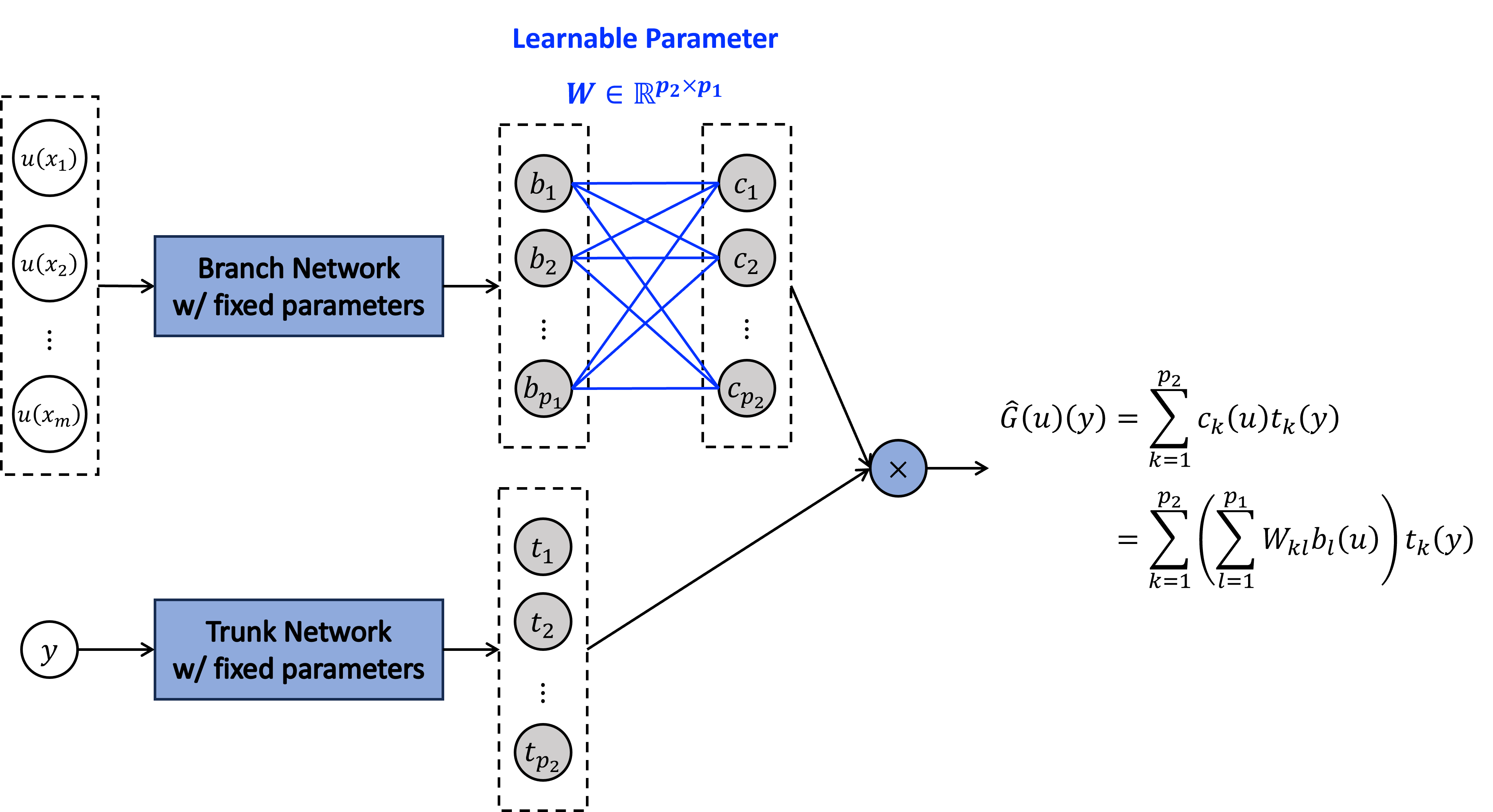}
  \caption{An illustration of the proposed ELM-DeepONets. The parameters of the branch and trunk networks are fixed during training, while an additional learnable parameter $W \in \mathbb{R}^{p_2\times p_1}$ is introduced to generate the output $G(u)(y)$.}
  \label{elm-deeponet}
\end{figure*}

Let \begin{equation}
    \widehat{G}(u) = \sum_{k=1}^{p_2} c_k(u)t_k(y), \nonumber
\end{equation} be the ELM-DeepONet output, where $t_k(y)$, for $k=1,2,\dots,p_2$ are the basis functions modeled by the trunk network with fixed parameters, and $p_2$ is a hyperparameter. To compute $c_k(u)$, we use another fully connected neural network with fixed parameters to generate $b_k(u)$ for $k=1,2,\dots,p_1$, and define $c_k(u)$ as: \begin{align}
    \begin{pmatrix}
        c_1(u) \\ c_2(u) \\ \vdots \\ c_{p_2}(u) 
    \end{pmatrix} &= \begin{pmatrix}
        W_{11} & W_{12} & \dots & W_{1p_1} \\
        W_{21} & W_{22} & \dots & W_{2p_1} \\
        \vdots & \vdots & \vdots & \vdots \\
        W_{p_2 1} & W_{p_2 2} & \dots & W_{p_2 p_1}
    \end{pmatrix}\begin{pmatrix}
        b_1(u) \\ b_2(u) \\ \vdots \\ b_{p_1}(u) 
    \end{pmatrix}\nonumber\\[5pt]
    &=: \mathbf{W} \begin{pmatrix}
        b_1(u) \\ b_2(u) \\ \vdots \\ b_{p_1}(u) 
    \end{pmatrix},\nonumber
\end{align} where $\mathbf{W}\in\mathbb{R}^{p_2\times p_1}$ is a learnable parameter. Consequently, the output of ELM-DeepONet can be expressed as:
\begin{align}
    \widehat{G}(u)(y)& = \sum_{k=1}^{p_2} c_k(u) t_k(y), \nonumber\\
    &= \sum_{k=1}^{p_2} \left( \sum_{l=1}^{p_1} W_{kl}b_l(u) \right)t_k(y), \nonumber
\end{align} where $p_1$ and $p_2$ are hyperparameters that will be discussed in detail later. In this novel framework, the only learnable parameter is $\mathbf{W}$, while all the parameters in $b_k(u)$ and $t_k(y)$ remain fixed. This design drastically reduces the number of trainable parameters in ELM-DeepONet to $p_1p_2$ which is significantly smaller than that of the vanilla DeepONet. The overall architecture is illustrated in \ref{elm-deeponet}.

Let $\mathcal{D} = \{(u_i, G(u_i)(y_j))\}_{i,j=1}^{N,M}$ be a supervised dataset for training ELM-DeepONet. Then the objective $$ \mathcal{L} =  \sum_{i=1}^{N}\mathcal{L}_i = \sum_{i=1}^N \sum_{j=1}^M \left(\widehat{G}(u_i)(y_j) - G(u_i)(y_j) \right)^2$$ can be expressed as:
\begin{equation}\label{leastsquare}
    \mathcal{L} = \Vert \mathbf{T} \mathbf{W} \mathbf{B} - \mathbf{G}\Vert_F^2,
\end{equation} where $\Vert \cdot \Vert_F$ denotes the Frobenius norm, and
\begin{align}
    \mathbf{T} &= \begin{pmatrix}
        t_1(y_1) & t_2(y_1) & \dots & t_{p_2}(y_1) \\
        t_1(y_2) & t_2(y_2) & \dots & t_{p_2}(y_2) \\
        \vdots & \vdots & \vdots & \vdots  \\
        t_1(y_M) & t_2(y_M) & \dots & t_{p_2}(y_M)
    \end{pmatrix} \in \mathbb{R}^{M\times p_2},\nonumber \\[5pt]
    \mathbf{W} &= \begin{pmatrix}
        W_{11} & W_{12} & \dots & W_{1p_1} \\
        W_{21} & W_{22} & \dots & W_{2p_1} \\
        \vdots & \vdots & \vdots & \vdots \\
        W_{p_2 1} & W_{p_2 2} & \dots & W_{p_2 p_1}
    \end{pmatrix} \in \mathbb{R}^{p_2\times p_1},\nonumber \\[5pt]
    \mathbf{B} &= \begin{pmatrix}
        b_1(u_1) & b_1(u_2) & \dots & b_1(u_N) \\
        b_2(u_1) & b_2(u_2) & \dots & b_2(u_N) \\
        \vdots & \vdots & \vdots & \vdots \\
        b_{p_1}(u_1) & b_{p_1}(u_2) & \dots &b_{p_1}(u_N)
    \end{pmatrix} \in \mathbb{R}^{p_1\times N}, \nonumber\\[5pt]
    \mathbf{G} &= \begin{pmatrix}
        G(u_1)(y_1) & G(u_2)(y_1) & \dots & G(u_N)(y_1) \\
        G(u_1)(y_2) & G(u_2)(y_2) & \dots & G(u_N)(y_2) \\
        \vdots & \vdots & \vdots & \vdots \\
        G(u_1)(y_M) & G(u_2)(y_M) & \dots & G(u_N)(y_M) \\
    \end{pmatrix} \in \mathbb{R}^{M\times N}. \nonumber
\end{align} Here, $\mathbf{T}$ represents the output of the trunk network, $\mathbf{W}$ is the learnable parameter, $\mathbf{B}$ represents the output of the branch network, and $\mathbf{G}$ is the label.

The objective in Equation \eqref{leastsquare} can be minimized by using the Moore-Penrose pseudoinverse as:
$$ (\mathbf{T}^\dagger \mathbf{T}) \mathbf{W} (\mathbf{B} \mathbf{B}^\dagger) = \mathbf{T}^\dagger \mathbf{G} \mathbf{B}^\dagger.$$  To solve the equation for $\mathbf{W}$, we assume that $\mathbf{T}$ and $\mathbf{B}$ are of full rank. By selecting $p_2$ such that $p_2\leq M$, then $\mathbf{T}^T \mathbf{T} \in \mathbb{R}^{p_2 \times p_2}$ is of full rank. This allows us to compute the left pseudoinverse of T as: $$\mathbf{T}^\dagger = (\mathbf{T}^T\mathbf{T})^{-1}\mathbf{T}^T.$$ Similarly, if $p_1$ is chosen such that $p_1 \leq N$, then the matrix $\mathbf{B}\mathbf{B}^T$ is of full rank and the right pseudoinverse of $\mathbf{B}$ can be computed as: $$\mathbf{B}^\dagger=\mathbf{B}^T(\mathbf{B}\mathbf{B}^T)^{-1}.$$ Thus, by setting $p_1 \leq N$, and $p_2 \leq M$, we obtain the solution of the objective in Equation \eqref{leastsquare} as:
$$ \widehat{\mathbf{W}} = \mathbf{T}^\dagger \mathbf{G} \mathbf{B}^\dagger.$$
Consequently, with the proposed ELM-DeepONet, training a DeepONet is reduced to solving a least square problem, which can be efficiently addressed by computing two pseudoinverses. 

The proposed ELM-DeepONet architecture offers notable flexibility. For instance, the branch network can utilize a Convolutional Neural Network (CNN) with fixed weights, a modification of the common approach in modern DeepONet architectures \cite{lu2021learning, cho2024physics}. Additionally, since the trunk network is responsible for generating global basis functions, it can be replaced with predefined fixed basis functions, such as $\{\sin(k\pi x)\}_{k=1}^{p_2}$ or $\{\cos(k\pi x)\}_{k=1}^{p_2}$, instead of employing a neural network. These variations are evaluated through numerical experiments to assess their effectiveness and computational efficiency.

\section{Numerical Results}\label{experiments}
We present the numerical results demonstrating the superior performance of the proposed method. All experiments were conducted using a single NVIDIA GeForce RTX 3090 GPU.
\subsection{Ordinary Differential Equations}

\begin{figure}[t]
  \centering
  \includegraphics[width=\linewidth]{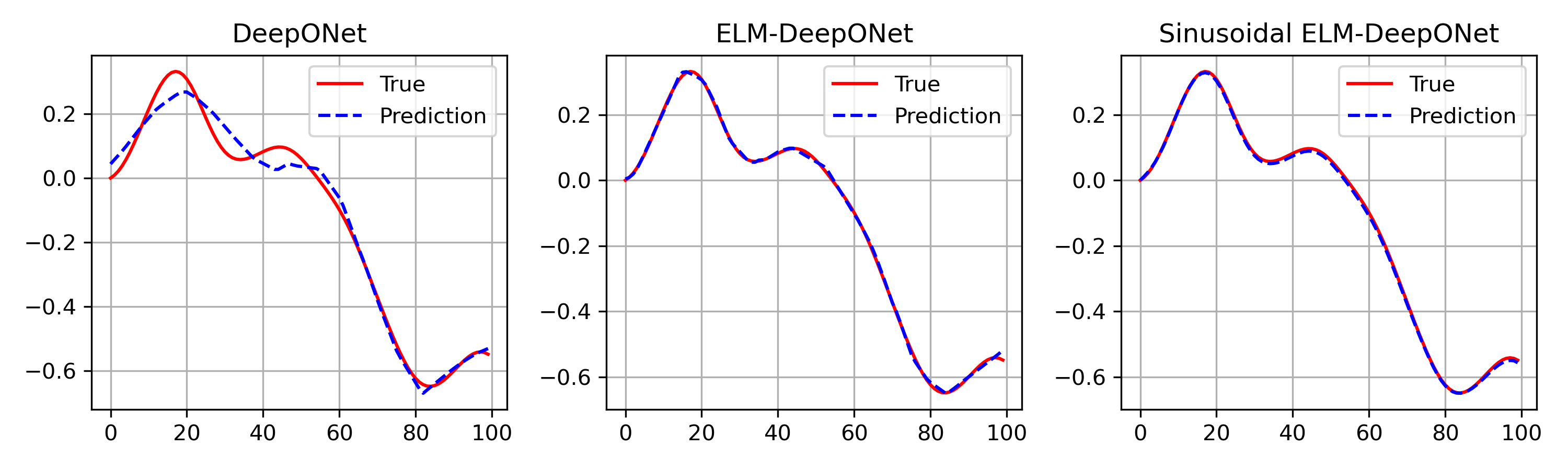}
  \caption{Numerical results for the antiderivative example on a single test sample demonstrate that DeepONet fails to capture the exact solution, whereas both ELM-DeepONets, one with fixed neural networks and the other with sinusoidal basis functions, accurately predict the true target function.}
  \label{Antiderivative}
\end{figure}

\subsubsection{Antiderivative}\label{sec:antiderivative} We consider learning the antiderivative operator: $$G: u(x) \mapsto s(x) = \int_0^x u(t)dt, $$ as studied in \cite{lu2021learning}. To generate the input functions, we sampled 2000 instances of $u_i$ from a Gaussian Random Field (GRF): $$u \sim \mathcal{G}(0, K_l(x_1, x_2)),$$  where the covariance kernel is defined by $K_l(x_1, x_2)=\exp(\frac{-\Vert x_1-x_2\Vert^2}{2l^2})$ with $l=0.1$. Using numerical integration, we created the dataset $\{(u_i, G(u_i)(y_j))\}_{i,j=1}^{N, M}$ with $N=2000$, $M=100$ where $y_j \in [0,1]$ represents the uniform collocation points. The dataset was split into 1000 training samples and 1000 testing samples.

As baseline models, we employed two DeepONets, M1 (Model 1) and M2 (Model 2), each utilizing fully connected neural networks with ReLU activation for both the branch and trunk networks. M1 features branch and trunk networks with architectures 1-64-64-64-1, while M2 employs larger networks with architectures 1-256-256-256-1. We trained the DeepONet by using Adam optimizer with a learning rate 1e-3 for  epochs \cite{kingma2014adam}. We fixed the branch network of the ELM-DeepONet with SLFNs with ReLU activation and the trunk network with 3-layer MLP consisting of $p_2$ hidden nodes in each layer. We also tested the sinusoidal basis function (Sinusoidal ELM-DeepONet) rather than the trunk net, such as $\{\sin(\frac{32k\pi}{2p_2}y)\}_{k=1}^{\frac{p_2}{2}} \cup \{\cos(\frac{32k\pi}{2p_2}y)\}_{k=1}^{\frac{p_2}{2}}$ . In this example, we set $p_1=N=1000$ and $p_2=M=100$. 

The results are summarized in Table \ref{Antiderivative_results}. As shown, training the ELM-DeepONet requires negligible time compared to the vanilla DeepONets, while the proposed methods achieve significantly lower relative test errors. Figure \ref{Antiderivative} illustrates the results on a representative test sample.

\begin{table}[ht]
\renewcommand{\arraystretch}{1.3}
\centering
\caption{The number of parameters, training time, and the relative $L^2$ test errors of the baseline DeepONets and the proposed methods for the antiderivative example. The proposed ELM-DeepONet achieves the best relative $L^2$ test error for the antiderivative example.}
\begin{tabular}{cccc}
\hline
Model                                                              & \# parameters & Training time (s) & Relative error \\ \hline
DeepONet M1                                                        & 23K           & 437               & 4.12\%               \\ \hline
DeepONet M2                                                        & 290K          & 495               & 7.19\%               \\ \hline
ELM-DeepONet                                                       & 20K           & \textbf{0.14}     & \textbf{2.12\%}      \\ \hline
\begin{tabular}[c]{@{}c@{}}Sinusoidal \\ ELM-DeepONet\end{tabular} & 20K           & \textbf{0.11}     & \textbf{2.45\%}      \\ \hline
\end{tabular}
\label{Antiderivative_results}
\end{table}

\subsubsection{Nonlinear ODE}
Next, we focus on learning the solution operator $G: u(x) \mapsto s(x)$ of a nonlinear ODE defined as:
\begin{equation}
    s'(x) = -0.1s^2(x) + u(x). \nonumber
\end{equation} the source function $u$ is sampled from the same GRF as described in Section \ref{sec:antiderivative}. Using numerical integration, we construct the dataset $\{(u_i, G(u_i)(y_j))\}_{i,j=1}^{N,M}$ with $N=2000$, $M=100$, and $y\in[0,1]$. We then compare the relative test errors of the proposed ELM-DeepONet and the Sinusoidal ELM-DeepONet to those of the baseline models M1 and M2. The results are summarized in Table \ref{Nonlinear_results}. 

\begin{table}[ht]
\renewcommand{\arraystretch}{1.3}
\centering
\caption{Summary of results for the nonlinear ODE example, showcasing the proposed methods' improved relative error and significantly reduced training time compared to the vanilla DeepONets.}
\begin{tabular}{cccc}
\hline
Model                                                              & \# parameters & Training time (s) & Relative error \\ \hline
DeepONet M1                                                        & 23K           & 451               & 4.87\%               \\ \hline
DeepONet M2                                                        & 290K          & 485               & 5.21\%               \\ \hline
ELM-DeepONet                                                       & 20K           & \textbf{0.14}     & \textbf{2.91\%}      \\ \hline
\begin{tabular}[c]{@{}c@{}}Sinusoidal \\ ELM-DeepONet\end{tabular} & 20K           & \textbf{0.11}     & \textbf{3.39\%}      \\ \hline
\end{tabular}
\label{Nonlinear_results}
\end{table}

\subsection{Sensitivity analysis}
The hyperparameters $p_1$ and $p_2$ play a critical role  in the performance of ELM-DeepONet. As previously discussed, it is important to set $p_1$ and $p_2$ to satisfy the condition \begin{equation}\label{condition}p_1\leq N,  p_2 \leq M,\end{equation} to ensure that the pseudoinverses $T^\dagger$ and $B^\dagger$ function as left and right inverses, respectively. However, we have observed that, in practice, significantly larger values of $p_1$ and $p_2$ which violate the condition in Equation \eqref{condition}, can still yield satisfactory results. We provide sensitivity analysis for the choice of parameters $p_1$ and $p_2$ through the antiderivative example.

Tables \ref{sensitivity} and \ref{sin_sensitivity} present the relative errors corresponding to various choices of $p_1$ and $p_2$. These relative errors were calculated over 10 trials and averaged to ensure statistical reliability. The results reveal several intriguing trends. First, as $p_2$ increases, the relative error consistently grows, reinforcing the empirical validity of the condition $p_2\leq M$. This observation aligns well with theoretical expectations and highlights the importance of carefully constraining $p_2$ for stable performance.

On the other hand, a contrasting pattern emerges for $p_1$; the relative error decreases as $p_1$ increases, even when $p_1>N$. This behavior deviates from the theoretical condition $p_1 \leq N$ suggesting that relaxing this constraint might improve performance in practical settings. These findings underline the relationship between the hyperparameters $p_1, p_2$ and the model's performance, offering valuable insights for their optimal selection in ELM-DeepONet.

\begin{table}[ht]
\centering
\renewcommand{\arraystretch}{1.3}
\caption{Test relative errors of ELM-DeepONet for the antiderivative example with different $p_1$ and $p_2$. We fix the branch network with an SLFN and the trunk network with a 3-layer MLP.}
\begin{tabular}{l|llllll}
\diagbox{$p_1$}{$p_2$} & 50    & 100   & 500   & 1000 & 5000& 10000 \\ \hline
50    & 15\%  & 15\%  & 16\%  & 17\%  & 27\% & 30\% \\
100    & 8.8\% & 8.8\%  & 9.2\% & 11\%  & 23\% & 26\% \\
500   & 3.3\% & 3.6\% & 5.5\% & 8.1\% & 22\% & 22\% \\
1000   & 2.2\% & 2.5\% & 4.1\% & 7.2\% & 22\% & 23\% \\
5000  & 1.7\% & 1.6\% & 2.3\% & 7.8\% & 20\% & 22\% \\
10000  & 1.8\% & 1.4\% & 3.4\% & 7.1\% & 22\% & 27\%
\end{tabular}\label{sensitivity}
\end{table}

\begin{table}[ht]
\centering
\renewcommand{\arraystretch}{1.3}
\caption{Test relative errors of Sinusoidal ELM-DeepONet for the antiderivative example with different $p_1$ and $p_2$. We fix the branch network with an SLFN and replace the trunk network with sinusoidal bases.}
\begin{tabular}{l|llllll}
\diagbox{$p_1$}{$p_2$} & 50    & 100   & 500   & 1000 & 5000& 10000 \\ \hline
50    & 15\%  & 16\%  & 16\%  & 14\%  & 15\% & 15\% \\
100    & 7.8\% & 7.9\%  & 7.5\% & 8.2\%  & 7.8\% & 8.0\% \\
500   & 5.5\% & 5.6\% & 5.3\% & 5.8\% & 5.6\% & 5.5\% \\
1000   & 2.7\% & 2.6\% & 2.7\% & 2.7\% & 2.6\% & 2.7\% \\
5000  & 0.6\% & 0.6\% & 0.6\% & 0.6\% & 0.7\% & 0.8\% \\
10000  & 0.4\% & 0.4\% & 0.4\% & 0.4\% & 0.4\% & 0.4\%
\end{tabular}\label{sin_sensitivity}
\end{table}

\subsection{Darcy Flow}
Next, we investigate the Darcy Flow, a 2D Partial Differential Equation (PDE) defined as:
\begin{align}
    \nabla \cdot (\kappa \nabla u) &= f, \text{ for } (x,y)\in\Omega := [0,1]^2, \nonumber\\
    u &= 0, \text{ for } (x,y)\in\partial\Omega, \nonumber
\end{align} where $\kappa$ represents the permeability and the source function is given by $f(x,y) = 1$.
The objective is to train the models to learn the solution operator, which maps the permeability field to the corresponding solution $$G: \kappa \mapsto u.$$ Here $\kappa$ is sampled from the GRF to generate diverse input scenarios. We use a $50\times50$ uniform mesh in $\Omega$ as collocation points. We create the dataset $\{(\kappa_i, G(\kappa_i)(x_j,y_j))\}_{i,j=1}^{N,M}$ with $N=2000$ and $M=2500$. 

As a baseline algorithm, we employed two DeepONets with different branch networks. One with a 3-layer MLP with each layer consisting of 128 nodes (DeepONet-MLP), and the other with a Convolutional Neural Network (CNN) with three convolutional layers followed by two fully connected layers (DeepONet-CNN). For the trunk network, we utilized a 3-layer Multilayer Perceptron (MLP), with each layer consisting of 128 nodes. We use ReLU as an activation function for all networks. We trained the network by using Adam optimizer for 10000 epochs with a learning rate 1e-3. 

\begin{figure}[ht]
  \centering
  \includegraphics[width=0.6\linewidth]{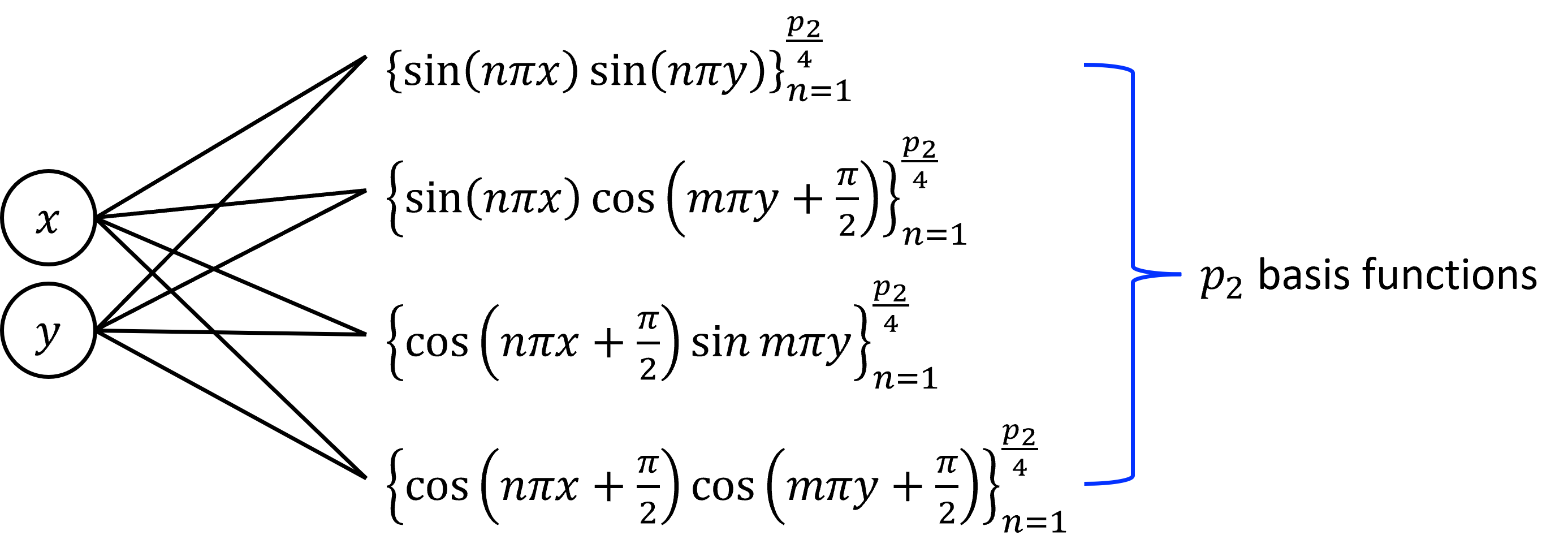}
  \caption{An illustration of sinusoidal basis functions for the Darcy Flow example. Fixed basis functions are used as an alternative to the trunk network. The solid lines are included solely for visual consistency with the trunk network and do not represent weights in this context.}
  \label{sinusoidal}
\end{figure}

We employed two ELM-DeepONets with MLP branch and CNN branch network, as the DeepONets, with all parameters kept fixed. In the Sinusoidal ELM-DeepONet, the trunk network was replaced with sinusoidal basis functions, specifically: $\{\sin(n\pi x)\sin(n\pi y), \sin(n\pi x)\cos(n\pi y + \pi/2), \cos(n\pi x + \pi/2)\sin(n\pi y), \cos(n\pi x + \pi/2)\cos(n\pi y + \pi/2)\},$ 
designed to enforce boundary conditions and leverage the advantages of harmonic representations, see Figure \ref{sinusoidal}. We choose the hyperparameters $p_1$ and $p_2$ by using a simple grid search. 

\begin{figure*}[t]
  \centering
  \includegraphics[width=1\linewidth]{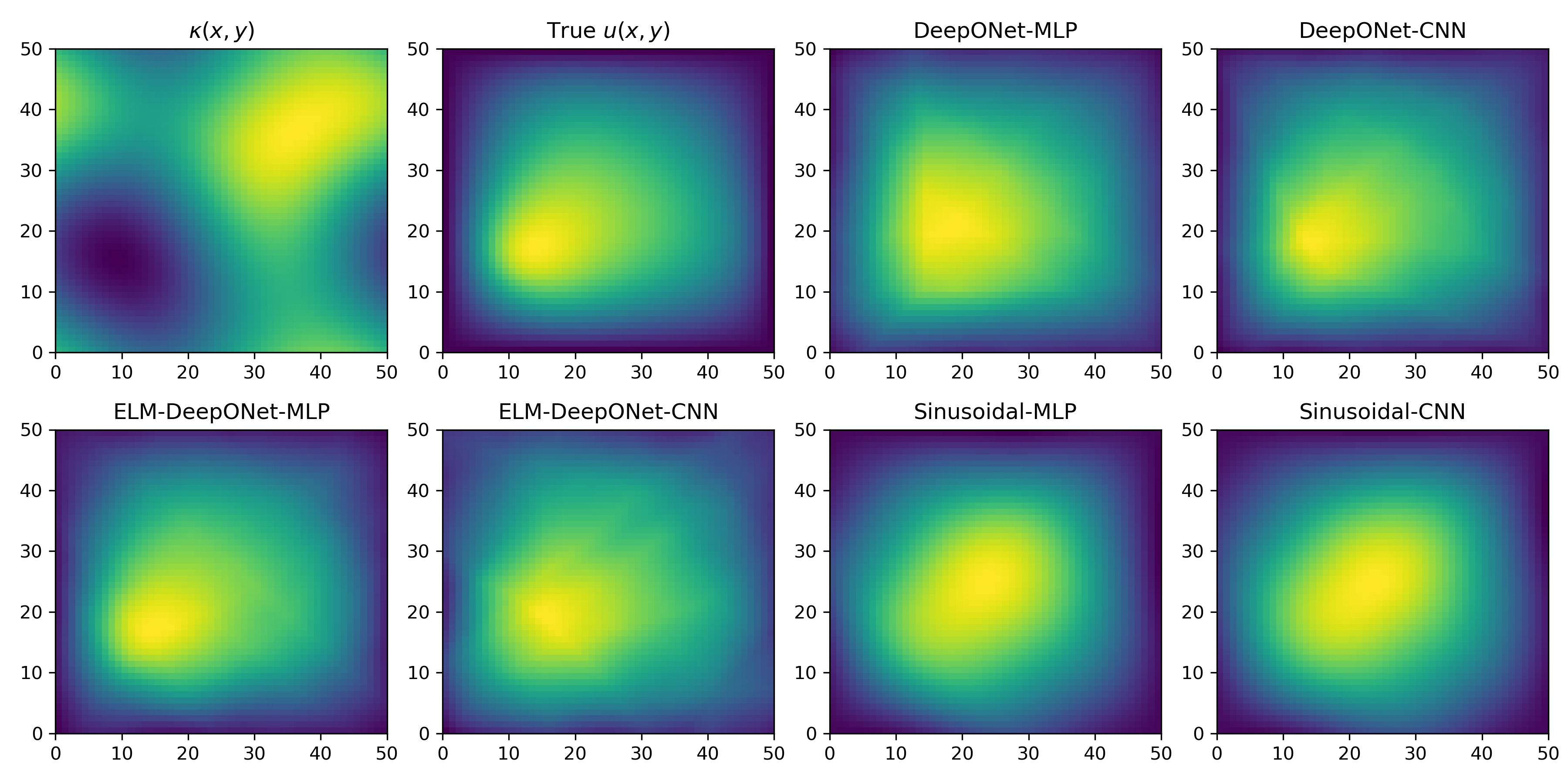}
  \caption{Results for a randomly selected test sample demonstrate that the ELM-DeepONet with an MLP branch network most accurately captures the solution, outperforming other models}
  \label{Darcy_result}
\end{figure*}

Table \ref{Darcy} showcases the superior performance of ELM-DeepONet compared to the vanilla DeepONet. Among the tested models, the ELM-DeepONet with an MLP branch network outperforms all others, achieving the best results in both relative error and training efficiency. The vanilla DeepONet with a CNN branch network ranks second, demonstrating competitive accuracy but requiring significantly more training time. The notable improvement of ELM-DeepONet with an MLP branch network highlights its ability to combine high accuracy with computational efficiency. Figure \ref{Darcy_result} visualizes the model predictions for a randomly selected test sample, further illustrating the effectiveness of the proposed approach.

\begin{table}[ht]
\renewcommand{\arraystretch}{1.3}
\centering
\caption{Summary of results for the Darcy Flow example: The proposed ELM-DeepONet with an MLP branch network has a significantly larger number of parameters compared to the DeepONets. However, it achieves much shorter training times while maintaining smaller relative errors.}
\begin{tabular}{ccccc}
\hline
\multicolumn{2}{c}{Model} & \# parameters & \begin{tabular}[c]{@{}c@{}}Training \\ time (s)\end{tabular}  & \begin{tabular}[c]{@{}c@{}}Relative \\ $L^2$ error\end{tabular} \\ \hline
\multirow{2}{*}{DeepONet} & MLP & 181K          & 1131              & 11.77\%              \\ \cline{2-5} 
 & CNN & 68K           & 1195              & 6.32\%               \\ \hline
\multirow{2}{*}{ELM-DeepONet} & MLP & 10M         & 3.10              & \textbf{5.65\%}               \\ \cline{2-5} 
 & CNN & 500K          & 3.02    & 6.80\%     \\ \hline
\multirow{2}{*}{\begin{tabular}[c]{@{}c@{}}Sinusoidal \\ ELM-DeepONet\end{tabular}} & MLP & 250K         & 0.99    &15.24\%     \\ \cline{2-5} 
 & CNN & 5M          & 3.13              & 15.01\%              \\ \cline{1-1} \hline
\end{tabular}\label{Darcy}
\end{table}

\subsection{Reaction Diffusion equation: An inverse source problem}
We consider an inverse source problem for a parabolic equation, as discussed in both the PINN and operator learning literature \cite{zhang2023stability, cho2024physics}. The equation is defined as: \begin{align}
    u_t &= Du_xx + ku^2 + s(x), &&(t,x)\in \Omega_T :=[0,1]\times\Omega,  \nonumber \\
    u(0,x) &= u_0(x), &&x\in \Omega,  \nonumber\\
    u(t,x) &= 0, &&(t,x) \in [0,1]\times\partial\Omega, \nonumber
\end{align} where $\Omega=[0,1]$. The objective is to learn the inverse operator $$G: u\vert_{\partial \Omega_T} \mapsto s,$$ which maps the boundary observations of $u$ to the source term $s(x)$. \cite{zhang2023stability} established that this problem is well-posed.

\begin{figure*}[t]
  \centering
  \includegraphics[width=1\linewidth]{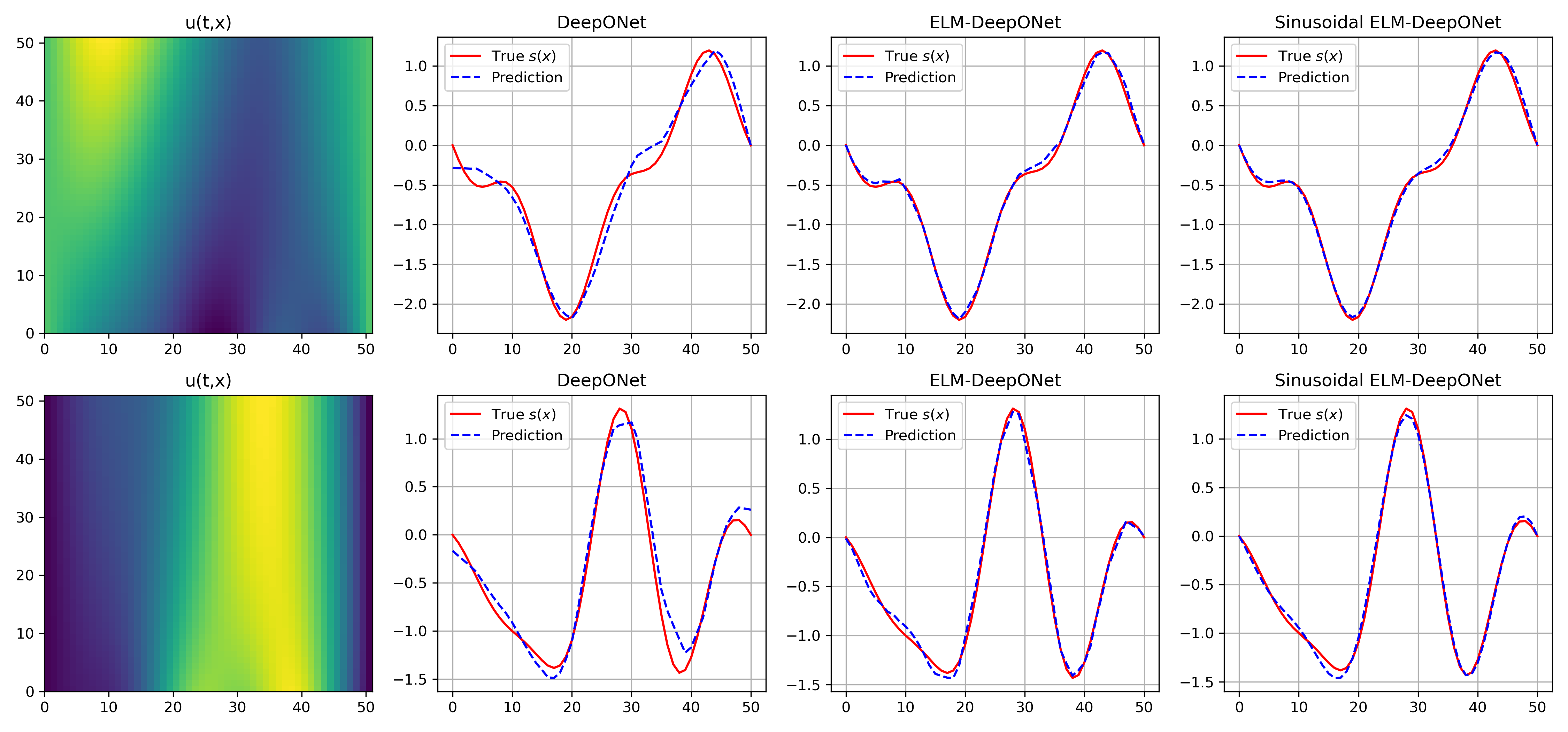}
  \caption{Results for two randomly selected test samples demonstrate that the proposed methods outperform the vanilla DeepONet.}
  \label{RD_result}
\end{figure*}

As a baseline algorithm, we employed a vanilla DeepONet with MLP-based branch and trunk networks, each consisting of three layers with 64 nodes per layer. For the ELM-DeepONet, we utilized an SLFN for the branch network and a three-layer MLP for the trunk network. The hyperparameters $p_1=10000$ and $p_2=50$ were selected using a grid search method. We summarized the result in Table \ref{RD_results}, and Figure \ref{RD_result} presents the results on two randomly selected test sample.

\begin{table}[ht]
\renewcommand{\arraystretch}{1.3}
\centering
\caption{Summary of results for the inverse source problem, showcasing the proposed methods' improved relative error and significantly reduced training time compared to the vanilla DeepONets.}
\begin{tabular}{cccc}
\hline
Model                                                              & \# parameters & Training time (s) & Relative error \\ \hline
DeepONet                                                         & 23K          & 104               & 5.04\%               \\ \hline
ELM-DeepONet                                                       & 500K           & \textbf{0.15}     & \textbf{2.77\%}      \\ \hline
\begin{tabular}[c]{@{}c@{}}Sinusoidal \\ ELM-DeepONet\end{tabular} & 500K           & \textbf{0.11}     & \textbf{2.52\%}      \\ \hline
\end{tabular}
\label{RD_results}
\end{table}

\section{Conclusion}\label{discussion}
We propose a novel training algorithm, ELM-DeepONet, to enable the efficient training of DeepONet by leveraging the least square formulation and computational efficiency of ELM. Through extensive validation, we have demonstrated that ELM-DeepONet outperforms vanilla DeepONet in both test error and training time. We believe that ELM-DeepONet represents a significant advancement for the operator learning community, addressing the challenge of high computational costs.

We leave several points for future work. First, the choice of $p_1$ and $p_2$ significantly impacts the performance of ELM-DeepONet. Empirically, we observed that larger values of $p_1$ and smaller values of $p_2$ generally lead to better performance. Notably, extremely large $p_1$, such as $p_1=10000$, performed well in practice, despite violating our theoretical insight that the pseudoinverse should act as a proper right inverse. A more thorough analysis of this phenomenon would be valuable for further advancing the framework. Second, recent studies have incorporated ELM into Physics-Informed Neural Networks (PINNs). Building on this trend, it is natural to consider extending ELM to Physics-Informed DeepONets as a promising direction for future work.


\bibliographystyle{unsrt}  
\bibliography{references}  






\end{document}